\pdfoutput=1
\documentclass[11pt]{article}

\usepackage[preprint]{acl}

\usepackage{dcolumn}
\newcolumntype{d}[1]{D{.}{.}{#1}}
\usepackage{times}
\usepackage{latexsym}
\usepackage{booktabs}
\usepackage{multirow}
\usepackage{amsmath} % For math formatting
\usepackage{amsfonts} % For math fonts
\usepackage{algorithm} % 加载 algorithm 宏包
\usepackage{algpseudocode} % 加载 algpseudocode 宏包，用于伪代码的排版
\usepackage{amsfonts} % For math fonts
\usepackage{amssymb} % For math symbols
\usepackage{subcaption}
\usepackage{pgfplots}
\pgfplotsset{compat=1.17}
\usepackage[T1]{fontenc}
\usepackage[utf8]{inputenc}
\usepackage{colortbl}
\usepackage{array}
\usepackage{booktabs}
\usepackage{multirow}
\usepackage{microtype}

\usepackage{inconsolata}

\usepackage{graphicx}

\title{HyperAdaLoRA: Accelerating LoRA Rank Allocation During Training via Hypernetworks without Sacrificing Performance}

\author{\bf Hao Zhang$^{1}$\thanks{These authors contribute equally to this work.}, Zhenjia Li$^{2}$\footnotemark[1], Runfeng Bao$^{3}$\footnotemark[1], Yifan Gao$^4$,  Xi Xiao$^5$,  Heng Zhang$^6$, \\ \bf Shuyang Zhang$^7$, Bo Huang$^2$,  Yuhang Wu$^8$, Tianyang Wang$^5$, Hao Xu$^{1}$\thanks{Corresponding author.}\\
\normalsize{}$^1$Harvard University,
\normalsize{}$^2$University of Chinese Academy of Sciences,
\normalsize{}$^3$Fudan University \\
\normalsize{}$^4$University of Chicago,
\normalsize{}$^5$University of Alabama at Birmingham,
\normalsize{}$^6$South China Normal University \\ 
\normalsize{}$^7$Shanghai Artificial Intelligence Laboratory,
\normalsize{}$^8$Shanghai University of Engineering Science }
\begin{document}
\maketitle

\begin{abstract}

Parameter-Efficient Fine-Tuning (PEFT), especially Low-Rank Adaptation (LoRA), has emerged as a promising approach to fine-tuning large language models(LLMs) while reducing computational and memory overhead. However, LoRA assumes a uniform rank \textit{r} for each incremental matrix, not accounting for the varying significance of weight matrices across different modules and layers. AdaLoRA leverages Singular Value Decomposition (SVD) to parameterize updates and employs pruning of singular values to introduce dynamic rank allocation, thereby enhancing adaptability. However, during the training process, it often encounters issues of slow convergence speed and high computational overhead. To address this issue, we propose HyperAdaLoRA, a novel framework that accelerates the convergence of AdaLoRA by leveraging a hypernetwork. Instead of directly optimizing the components of Singular Value Decomposition $(P, \Lambda, Q)$, HyperAdaLoRA employs a hypernetwork based on attention mechanisms to dynamically generate these parameters. By pruning the outputs of the hypernetwork that generates the singular values, dynamic rank allocation is achieved. Comprehensive experiments on various datasets and models demonstrate that our method achieves faster convergence without sacrificing performance. Additionally, further extension experiments on other LoRA-based approaches validate the broad applicability of our method.

\end{abstract}


\section{Introduction}

Parameter-efficient fine-tuning (PEFT) has emerged as a practical solution for adapting large language models (LLMs) to downstream tasks by updating a small subset of parameters, thereby reducing computational and memory overhead~\cite{li2021,lester2021,zaken2022,zhang2025pdtrim,zhang2025trimtokenator,houlsby2019}. A prominent PEFT method, Low-Rank Adaptation (LoRA) \cite{hu2022lora}, is particularly notable for introducing trainable low-rank matrices into pre-trained weights during fine-tuning, reparameterizing weight updates as:
\begin{equation}
W = W^{(0)} + \Delta W = W^{(0)} + BA
\end{equation}
where \( W^{(0)}, \Delta W \in \mathbb{R}^{d_1 \times d_2} \), \( A \in \mathbb{R}^{r \times d_2} \) and \( B \in \mathbb{R}^{d_1 \times r} \) with \( r \ll \{d_1, d_2\} \). During fine-tuning, only matrices \( B \) and \( A \) are updated, substantially reducing the number of trainable parameters. 
However, LoRA assigns a uniform rank across all layers, neglecting the varying functional importance of different components, potentially limiting performance in deep or heterogeneous models \cite{hu2023structure,zhang2023increlora}.

\begin{figure*}[ht]
    \centering
    \includegraphics[width=0.7\textwidth]{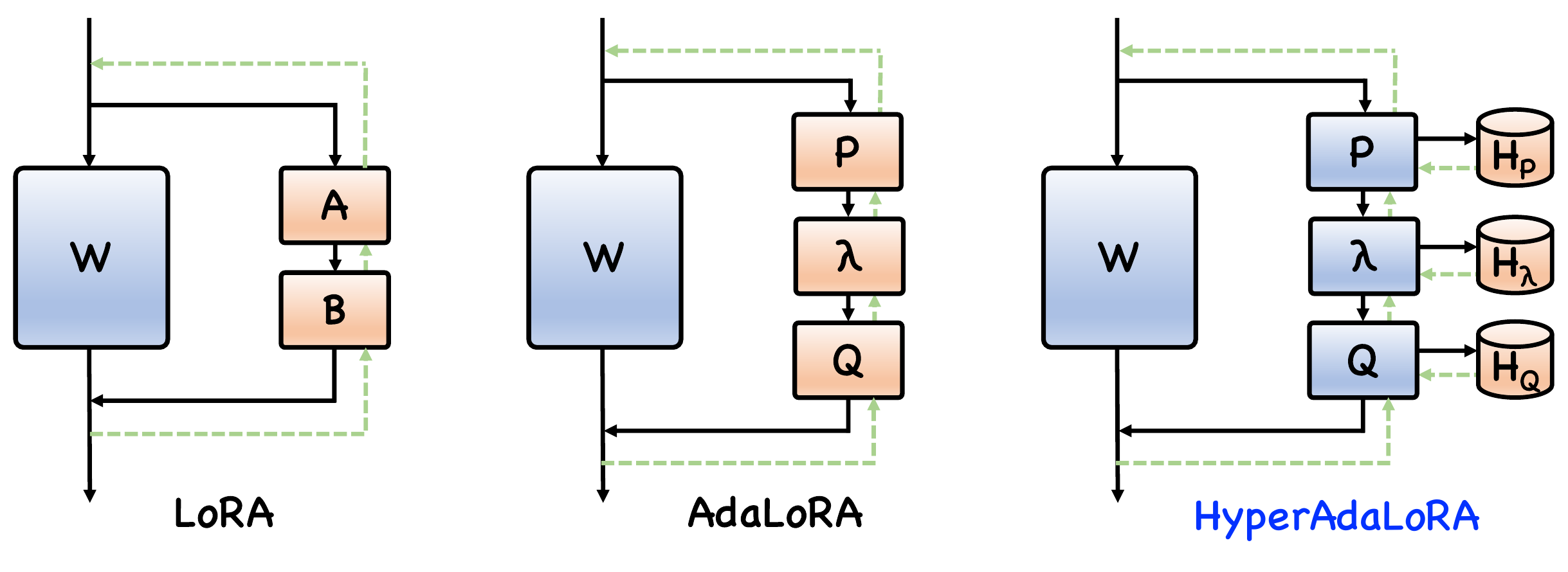}
    \caption{
        Comparison of LoRA, AdaLoRA, and HyperAdaLoRA frameworks, with black solid lines representing the forward process and green dashed lines indicating backpropagation (gradient flow). LoRA applies fixed-rank low-rank adaptations (\( A, B \)). AdaLoRA introduces dynamic rank allocation via Singular Value Decomposition (SVD) and singular value pruning (\( P, \lambda, Q \)). HyperAdaLoRA leverages a hypernetwork to dynamically generate SVD components (\( H_P, H_\lambda, H_Q \)), accelerating convergence and enhancing computational efficiency.
    }
    \label{figure1}
\end{figure*}

Dynamic rank allocation methods have been introduced to tackle this challenge by adaptively assign different ranks \( r \) to various modules or layers. There are three main strategies:  
1) Singular value decomposition (SVD) methods \cite{zhang2023adalora,hu2023structure,zhang2023increlora} decompose matrices into singular values and vectors, effectively capturing key components. However, the decomposition process is computationally intensive, with a time complexity of \( O(n^3) \), and requires additional memory to store the singular values and vectors.
2) Single-rank decomposition (SRD) methods \cite{mao2024dora,zhang2024autolora,liu2024alora} decompose matrices into rank-1 components, enabling more granular rank allocation. Despite this flexibility, identifying and pruning rank-1 components necessitates multi-stage training, increasing algorithmic complexity. The iterative selection process can also introduce instability, particularly when essential components are mistakenly pruned.
3) Rank sampling methods \cite{valipour2022dylora} dynamically allocates ranks during training by sampling from a range of ranks, offering post-training flexibility. However, the stochastic nature of this approach introduces gradient noise, potentially destabilizing convergence. 

Hypernetworks \cite{ha2016hypernetworks} is a meta-model that generates parameters for a target network, decoupling parameter generation from model architecture. Some fine-tuning methods \cite{ortiz2024hyperloader,lianghart,li2025hyperlora} leverage hypernetworks to enhance performance on both single-task and multi-task settings. By producing task-specific parameters based on contextual inputs, hypernetworks enable dynamic adaptation without explicit iterative optimization, capturing intricate parameter adaptation patterns in real-time. Hypernetworks facilitate adaptive parameter optimization by modulating both the update direction and magnitude in response to the current model state \cite{lorraine2018stochastic}. This dynamic modulation fosters efficient navigation through high-dimensional parameter spaces, accelerating convergence while maintaining model expressiveness \cite{kirsch2018modular,shi2022layerconnect}.  

In this paper, we introduce HyperAdaLoRA, a novel framework that leverages hypernetworks \cite{ha2016hypernetworks} to achieve a significant improvement in convergence speed through parameter generation. Unlike traditional methods that directly train the incremental matrices \(P\), \(\Lambda\), and \(Q\), HyperAdaLoRA employs task-specific hypernetworks to generate these matrices. Architecturally, our hypernetworks are based on a specific attention layer of BERT~\cite{devlin2019bert}, which enables them to capture the complex dependencies among parameters. Specifically, each hypernetwork takes the current state of \(P\), \(\Lambda\), or \(Q\) as input and outputs their updated versions. Dynamic rank allocation is realized by pruning the output of the hypernetwork that generates \(\Lambda\). The training objective of HyperAdaLoRA is to minimize the discrepancy between the parameters generated by the hypernetworks and the ideal parameters. Extensive experiments demonstrate that our method achieves faster convergence while maintaining accuracy. Additionally, further extension experiments on other LoRA-based (SRD and rank-sampling) approaches validate the broad applicability of our method.

The main contributions of our paper can be summarized as follows:
\begin{itemize}
    \item We introduce HyperAdaLoRA, a pioneering framework that utilizes hypernetworks to achieve substantial acceleration in convergence speed through advanced parameter generation.
    \item We employ attention based hypernetworks to capture the complex dependencies among parameters and accurately perform parameter updates during the training process.   
    \item We conducted extensive experiments showing that HyperAdaLoRA achieves faster convergence without sacrificing accuracy, and validated its broad applicability on other LoRA-based methods.
\end{itemize}

\section{Related Work}
\subsection{SVD-based Fine-tuning Method} 
\label{sec:svd_dynamic}

SVD-Based methods parameterize LoRA’s low-rank update matrix  $\Delta W$ in a singular value decomposition form (e.g. splitting into $P \Lambda Q$) to dynamically adjust effective rank. 
The mathematical representation is as follows:
\begin{equation}
W = W^{(0)} + \Delta W = W^{(0)} + P \Lambda Q
\end{equation}
where \( P \in \mathbb{R}^{d_1 \times r} \) and \( Q \in \mathbb{R}^{r \times d_2} \) represent the left/right singular vectors, and the diagonal matrix \( \Lambda \in \mathbb{R}^{r \times r} \) contains the singular values \( \{\lambda_i\}_{1 \leq i \leq r} \) with \( r \ll \min(d_1, d_2) \).
For example, AdaLoRA \cite{zhang2023adalora} prunes less important singular values based on a sensitivity-derived importance score during training. SaLoRA \cite{hu2023structure} adaptively adjusts the rank by identifying and suppressing less informative singular components, optimizing parameter efficiency across layers. IncreLoRA \cite{zhang2023increlora} incrementally increases the rank during training, starting with a minimal rank and expanding as needed, balancing early training stability with later-stage expressiveness. These methods effectively capture principal components but incur \( O(n^3) \) complexity and substantial memory overhead, impacting scalability and training stability, particularly with dynamic rank adjustment.

\subsection{SRD-based Fine-tuning Method}
SRD-based methods decompose the LoRA update \( \Delta W = \sum_{i=1}^r u_i v_i^T \) into a series of rank-1 matrices, where each component \( u_i v_i^T \) represents a distinct direction in the parameter space. This decomposition allows the model to assess and adjust each rank-1 update independently.
AutoLoRA \cite{zhang2024autolora} uses a meta-learning scheme to determine which rank-1 slices to retain or prune, while ALoRA \cite{liu2024alora} trains a 'super-network' and reallocates ranks based on importance. SoRA \cite{ding2023sparse} applies sparsity penalties to zero out less impactful components, and DoRA \cite{liu2024dora} adjusts only the direction component, assigning a learnable scalar for each weight. 
However, SRD methods face limitations such as increased algorithmic complexity from multi-stage training and additional optimization for rank-1 component selection. Abrupt pruning can destabilize training, while iterative selection adds computational overhead and heightens sensitivity to hyperparameter tuning.

\subsection{Rank Sampling-based Fine-tuning Method}
Rank-sampling methods treat the LoRA rank as a random variable during training. DyLoRA \cite{valipour2022dylora} implements this by sampling a truncation level \( b \leq R \) in each iteration, zeroing out the bottom \( R - b \) components and enabling the model to operate across multiple ranks without retraining. This approach eliminates the need for exhaustive rank search while also serving as a regularizer, concentrating key features in top components to potentially enhance generalization. However, training across multiple ranks can dilute performance at specific ranks compared to fixed-rank LoRA, and dynamic masking introduces slight computational overhead, potentially requiring additional training epochs to converge. Additionally, we note that some works \cite{li2025beyond,li2025beyond2} also introduce the concepts of non-zero initialization dynamics and token-wise input–output projections.

\section{Method}

\subsection{Preliminary} 
Hypernetworks \cite{ha2016hypernetworks} are a type of neural network architecture used to generate the weights of another neural network (the target network). This can be expressed using the following formula:
\begin{equation}
\Theta = H(C; \Phi)
\end{equation}
where \( \Theta \) represents the weights of the target network, \( H \) denotes the hypernetwork, \( C \) is the context vector input to the hypernetwork and \( \Phi \) corresponds to the weights of the hypernetwork itself.

The input to the hypernetwork \cite{ha2016hypernetworks} can be any information related to the target network, such as the input data of the target network, task requirements, or other contextual information. By conditioning parameter generation on these inputs, the hypernetwork can dynamically generate different parameters to adapt to different tasks or environments. This approach mitigates the need for learning a full set of parameters, thereby reducing model complexity and promoting generalization.

In neural architecture search (NAS), hypernetworks \cite{ha2016hypernetworks} can generate parameters for multiple sub-networks, thereby efficiently exploring different network architectures. By simultaneously training the hypernetwork and the sub-networks, the performance of a large number of candidate architectures can be evaluated in a relatively short time. This allows for the rapid screening of network structures with better convergence performance. This efficient architecture exploration method helps to find more optimal model architectures, thereby accelerating the overall training and convergence process. We present a more comprehensive theoretical analysis that elucidates how hypernetworks accelerate convergence (Appendix \ref{Comprehensive Analysis of Hypernetworks for Accelerating Convergence}).

\subsection{Hypernetworks Accelerate Convergence}
To accelerate the convergence of AdaLoRA, we propose a novel training strategy: employing a hypernetwork to dynamically generate the $P \Lambda Q$ parameters during training, rather than relying on traditional backpropagation for their updates. Specifically, at the beginning of training, we initialize the $P{\Lambda}Q$ parameters using a normal distribution. As training progresses, the parameters of the hypernetwork are continuously updated through backpropagation, thereby optimizing the generation process of the $P{\Lambda}Q$ parameters. In this process, the $P{\Lambda}Q$ parameters serve merely as intermediate results, with their ultimate goal being the optimized generation via the hypernetwork to achieve faster convergence. The design of the hypernetwork is crucial to this strategy. It takes the parameters before the update as input and, after a series of complex computations and optimization operations, outputs the updated parameters. To conserve computational resources and memory usage, we employ the same hypernetwork for the same parameters across different parameters. For example, we use a single hypernetwork to generate the updated values for the $P$ parameters of all matrix weights. This design not only improves resource efficiency but also ensures consistency and stability in parameter updates. In the $i$-th iteration, the update process of $P{\Lambda}Q$ can be specifically represented as follows:
\begin{align}
P_{i+1} &= \mathcal{H}_P(P_{i}; \Phi_P) \\
\Lambda_{i+1} &= \mathcal{H}_\Lambda(\Lambda_{i}; \Phi_\Lambda) \\
Q_{i+1} &= \mathcal{H}_Q(Q_{i}; \Phi_Q)
\end{align}
where $\mathcal{H}_P$, $\mathcal{H}_{\Lambda}$, and $\mathcal{H}_Q$ represent the hypernetworks used to update the parameters $P$, $\Lambda$, and $Q$, respectively. $\Phi_P$, $\Phi_{\Lambda}$, and $\Phi_Q$ represent the parameters of these hypernetworks. $P_i$, $\Lambda_i$, and $Q_i$ represent the parameters before the $i$-th iteration update, while $P_{i+1}$, $\Lambda_{i+1}$, and $Q_{i+1}$ represent the parameters after the update.

The process of a hypernetwork generating updated parameters by taking model parameters as input is essentially a dynamic parameter generation process. From a mathematical perspective, this can be likened to a nonlinear transformation within the parameter space, aimed at more efficiently approximating the optimal parameters. From the standpoint of optimization theory, such personalized updates can more effectively explore the parameter space to identify better solutions. Traditional optimization methods, such as gradient descent, follow fixed rules for parameter updates. In contrast, a hypernetwork can dynamically adjust the direction and magnitude of updates based on the current state of the parameters. This flexibility enables it to better adapt to complex loss landscapes and accelerate convergence. As a result, the hypernetwork can more precisely adjust the model's state in each iteration, thereby reducing the number of iterations required to achieve convergence.

During the initial training phases, the hypernetwork designed for $\Lambda$ generates a full-rank diagonal matrix. To facilitate adaptive budget allocation, we implement an iterative singular value pruning strategy based on magnitude thresholds. At each interval $\Delta T$, the $k$ smallest singular values within $\Lambda$ are set to zero, effectively reducing the rank of the incremental update $\Delta$. Subsequently, the hypernetworks adjust the patterns they generate to compensate for the pruned dimensions through a gradient-driven process of plasticity.

The loss function integrates task-specific objectives with orthogonality regularization, formulated as:
\begin{equation}
\mathcal{L} = \mathcal{L}_{\text{task}} + \gamma (\|P^\top P - I\|_F^2 + \|QQ^\top - I\|_F^2)
\end{equation}
where $\gamma > 0$ denotes the regularization coefficient. This formulation ensures that the generated matrices $P$ and $Q$ approximate orthogonal transformations while maintaining compatibility with downstream tasks. We additionally provide details on the parameter matrix update process (Appendix \ref{Processing Details of Parameter Matrices}) and the pruning strategy (Appendix \ref{Implementation Details of the Pruning Strategy}).

\subsection{Attention Driven Parameter Interaction}
We adopt a BERT layer as the architecture of the hypernetwork and leverage the self-attention mechanism to capture the dependencies among parameters. Specifically, the interaction between the query and key in the attention mechanism mimics the associations between elements in the parameter matrix. This enables the hypernetwork to generate context-aware updates that preserve the structural patterns of the parameters. For any parameter \( p_i \), its updated output takes into account all parameters in the matrix, as shown in the following equation:
\begin{equation}
p_{i+1} = \sum_{j=1}^{N} \text{Softmax}\left(\frac{Q_i K_j^T}{\sqrt{d}}\right) V_j
\end{equation}
where \( p_{i+1} \) represents the updated value of parameter \( p_i \), \( Q_i \) is the query vector associated with \( p_i \), \( K_j \) is the key vector associated with parameter \( p_j \), and \( V_j \) is the value vector corresponding to parameter \( p_j \). The total number of parameters in the matrix is denoted by \( N \), and \( d \) represents the dimensionality of the query and key vectors. This formulation allows the hypernetwork to dynamically compute the updated value of \( p_i \) by aggregating information from all parameters in the matrix, capturing their interdependencies and preserving the structural patterns of the parameters.

\begin{table*}[htbp]
\centering
\small
\begin{tabular}{cccccc}
\toprule
\textbf{Model} & \textbf{Method} & \multicolumn{2}{c}{\textbf{Stanford Alpaca}} & \multicolumn{2}{c}{\textbf{Magpie}}\\
\cmidrule(lr){3-4} \cmidrule(lr){5-6}
 & &  \textbf{BLEU-4} & \textbf{ROUGE-1} & \textbf{BLEU-4} & \textbf{ROUGE-1} \\
\midrule
\multirow{2}{*}{LLaMA3.1-8B} & AdaLoRA      & 55.06 & 58.51 & 70.69 & 56.76 \\
&  HyperAdaLoRA (ours) & \textbf{ 55.10} &  \textbf{58.58} &  \textbf{70.73} & \textbf{ 56.78}  \\

\midrule

\multirow{2}{*}{Qwen2.5-7B}   & AdaLoRA      & 6.79 & 20.17 & 56.21 & 49.43 \\
& HyperAdaLoRA (ours) & \textbf{6.79} &  \textbf{20.19} &  \textbf{56.22} &  \textbf{49.43} \\
\bottomrule
\end{tabular}

\caption{Performance comparison between HyperAdaLoRA and AdaLoRA on NLG tasks using LLaMA3.1-8B and Qwen2.5-7B as backbones. The reported metrics include BLEU-4 and ROUGE-1.}
\label{tab:nlg_performance}
\end{table*}

\begin{figure*}[ht]
\centering
% \begin{subfigure}[t]{0.32\textwidth}
%     \centering
%     \includegraphics[width=\textwidth]{figures/MNLI_RoBERTa-base.png}
%     \caption{MNLI / RoBERTa-base}
%     \label{fig:MNLI_RoBERTa}
% \end{subfigure}
% \hfill
\begin{subfigure}[t]{0.23\textwidth}
    \centering
    \includegraphics[width=\textwidth]{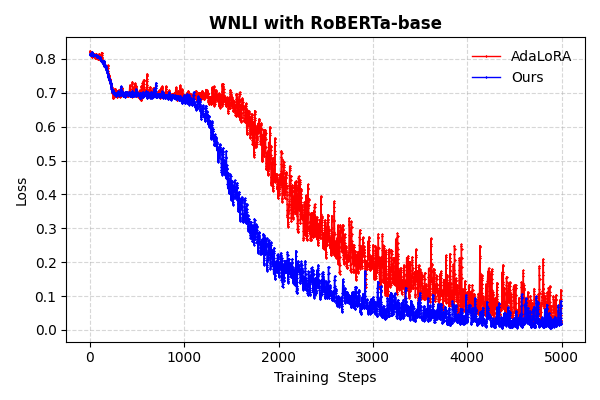}
    \caption{WNLI/RoBERTa-base}
    \label{fig:WNLI_RoBERTa}
\end{subfigure}
\hfill
\begin{subfigure}[t]{0.23\textwidth}
    \centering
    \includegraphics[width=\textwidth]{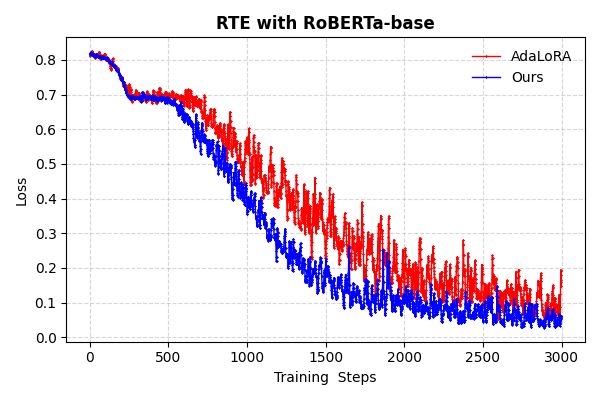}
    \caption{RTE/RoBERTa-base}
    \label{fig:RTE_RoBERTa}
\end{subfigure}
% \begin{subfigure}[t]{0.32\textwidth}
%     \centering
%     \includegraphics[width=\textwidth]{figures/MNLI_DeBERTa-v3-base.png}
%     \caption{MNLI / DeBERTa-v3-base}
%     \label{fig:MNLI_DeBERTa}
% \end{subfigure}
% \hfill
\begin{subfigure}[t]{0.23\textwidth}
    \centering
    \includegraphics[width=\textwidth]{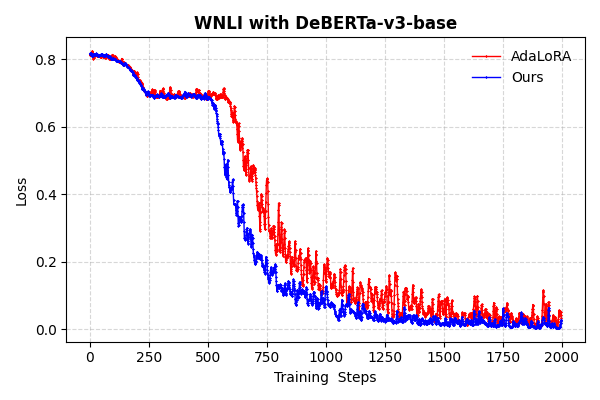}
    \caption{WNLI/DeBERTa-v3-base}
    \label{fig:WNLI_DeBERTa}
\end{subfigure}
\hfill
\begin{subfigure}[t]{0.23\textwidth}
    \centering
    \includegraphics[width=\textwidth]{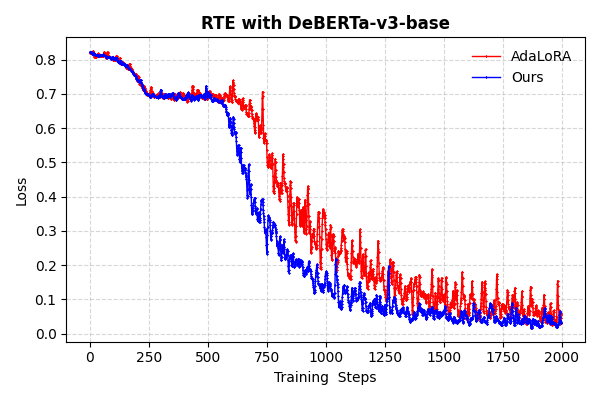}
    \caption{RTE/DeBERTa-v3-base}
    \label{fig:RTE_DeBERTa}
\end{subfigure}
\caption{Comparison of training loss convergence between HyperAdaLoRA and AdaLoRA on natural language understanding tasks. The rows correspond to three natural language understanding tasks: RTE and WNLI. The columns represent two pretrained language models: RoBERTa-base and DeBERTa-v3-base.}
\label{fig:main_convergence}
\end{figure*}

\section{Experiments}
\subsection{Experimental Setup} 
\subsubsection{Benchmarks} We conduct a comprehensive evaluation of our method, covering a wide range of tasks in both natural language understanding (NLU) and natural language generation (NLG). In the realm of natural language understanding, our method is tested on challenging tasks from the GLUE benchmark \cite{wang2018glue}: RTE \cite{wang2018glue} and WNLI \cite{wang2018glue}. These tasks represent large scale entailment classification, small-scale binary entailment classification, and coreference resolution presented in the form of binary entailment classification, respectively. In the natural language generation domain, we assess our method using three widely recognized datasets: Stanford Alpaca \cite{alpaca}, Magpie-Pro-300K-Filtered \cite{xu2024magpie}, and OpenPlatypus \cite{platypus2023}. In the following text, we sometimes abbreviate Magpie-Pro-300K-Filtered as Magpie.To further demonstrate the generalizability of our approach, we conduct additional evaluations on the GSM8K \cite{cobbe2021training} and HumanEval \cite{chen2021evaluating} benchmarks.

\subsubsection{Models} We use two prominent pretrained language models for NLU: RoBERTa-base \cite{liu2019roberta}, which is renowned for its strong performance across a wide range of NLU tasks, and DeBERTa-v3-base \cite{he2021debertav3}, an enhanced version that incorporates advanced pretraining techniques. For NLG, we employ two models: LLaMA3.1-8B \cite{grattafiori2024llama}, a powerful 8 billion parameter model optimized for high-quality text generation, and Qwen2.5-7B \cite{yang2024qwen2}, a model that demonstrates exceptional performance in various NLG tasks. We further evaluate our method on the larger Qwen2.5-14B \cite{yang2024qwen2}.

\subsubsection{Baselines} Our primary baseline for comparison is AdaLoRA \cite{zhang2023adalora}. AdaLoRA uses \( P \Lambda Q \) as trainable parameters that are dynamically updated during training. It allocates parameter budgets by parameterizing updates in an SVD form and prunes singular values based on importance scores during training. To further demonstrate the broad applicability of our method, we additionally conduct experiments by integrating it with LoRA \cite{hu2022lora}, DoRA (SRD-based) \cite{liu2024dora}, and DyLoRA (rank-sampling based) \cite{valipour2022dylora}.

\subsubsection{Implementation Details} Our experiments are conducted using the PyTorch framework \cite{paszke2019pytorch} and the Hugging Face Transformers library \cite{wolf2020transformers}, running on a cluster equipped with NVIDIA A100 40GB GPUs. In our experiments, we set the rank \( r \) to 3 and the orthogonality regularization coefficient \( \gamma \) to 0.1. We use the Adam optimizer \cite{kingma2014adam} with a learning rate of \( 1 \times 10^{-5} \) and a batch size of 64 for training. For the comparative experiments, we keep the hyperparameters for the base model fine-tuning consistent across different methods. We use ROUGE-1 and BLEU-4 as the NLG evaluation metrics. The hypernetwork is implemented as a single layer of TinyBERT, featuring a hidden dimension of 312 and 12 attention heads, with approximately 11.88 million parameters. For the CNN baseline, we adopt the ResNet-18 architecture, comprising 18 layers with hidden dimensions of 64, 128, and 256, totaling around 11.7 million parameters. Additionally, the MLP baseline consists of 15 layers with hidden dimensions of 256, 512, and 1024, amounting to approximately 11.4 million parameters. More training details can be found in Appendix \ref{Training Details}.

\subsection{NLG Task Results}
\subsubsection{Performance Comparison} We first compare the final generation quality of HyperAdaLoRA and AdaLoRA after fine-tuning the LLaMA3.1-8B and Qwen2.5-7B models on the Stanford Alpaca and Magpie datasets. The results are summarized in Table \ref{tab:nlg_performance} using BLEU-4 and ROUGE-1 scores. The results in Table \ref{tab:nlg_performance} indicate that HyperAdaLoRA does not exhibit any performance degradation compared to AdaLoRA. In most configurations, the scores of the two models are very close, with HyperAdaLoRA occasionally showing a slight edge. This demonstrates that the significant improvements in convergence speed do not negatively affect the final quality of the generated outputs.

\subsubsection{Training Efficiency} We further evaluate the effectiveness of HyperAdaLoRA in NLG tasks. We finetune the LLaMA3.1-8B and Qwen2.5-7B models on three instruction-following datasets: Stanford Alpaca, Magpie-Pro-300K-Filtered, and OpenPlatypus. Table \ref{tab:comparison} presents a comparison of the total training time. In all configurations, HyperAdaLoRA achieves shorter training times than AdaLoRA. This reduction is evident across datasets of varying sizes, from the large Magpie to the smaller Alpaca and OpenPlatypus datasets, consistent with the accelerated convergence. For instance, when fine-tuning LLaMA3.1-8B on the Stanford Alpaca dataset, HyperAdaLoRA takes 7250 seconds, compared to 8125 seconds for AdaLoRA. Similarly, when fine-tuning Qwen2.5-7B on the large Magpie-Pro dataset, HyperAdaLoRA has a training time of 14250 seconds, while AdaLoRA requires 15000 seconds. This consistent time advantage highlights the efficiency gains brought by hypernetwork based parameter generation. By more rapidly reaching effective parameter states, HyperAdaLoRA significantly reduces the total training duration needed for adaptation to these NLG tasks.

\subsection{NLU Task Results}
We compare the convergence speed of HyperAdaLoRA (which employs a BERT layered hypernetwork) with that of the baseline AdaLoRA. Figure \ref{fig:main_convergence} illustrates the training loss curves of these two methods on the  RTE and WNLI datasets, using RoBERTa-base and DeBERTa-v3-base as backbone models. Across all experimental settings, HyperAdaLoRA consistently converges significantly faster than AdaLoRA. The loss curves of HyperAdaLoRA drop more steeply in the early stages of training and reach a lower loss plateau earlier in the training process. Our method achieves convergence with fewer training steps. As shown in Table \ref{tab:efficiency}, it also incurs lower per-step latency, further demonstrating its advantage in fine-tuning efficiency.

\subsection{Analysis of Method Generalizability}
\subsubsection{Convergence Time on Other LoRA-Based Methods} To demonstrate this generality, we extend it beyond AdaLoRA to several representative methods, including LoRA, DoRA and DyLoRA. We evaluate these variants on LLaMA3.1-8B with a batch size of 8 using the Stanford Alpaca dataset. As shown in Table \ref{additional_time}, our method consistently accelerates convergence across all settings. We provide the details of hypernetworks for LoRA, DoRA, and DyLoRA in Appendix~\ref{appendix::detailsof}. Additional results are provided in Appendix~\ref{app:more_results}.

% \begin{table*}[ht]
%     \centering
%     \begin{tabular}{@{}ccccccc@{}}
%         \toprule
%         Method & LoRA & LoRA + Hyper & DoRA & DoRA + Hyper & DyLoRA & DyLoRA + Hyper \\
%         \midrule
%         Training Time (s) & 6893 & 5958 & 7525 & 6267 & 7129 & 6132 \\
%         \bottomrule
%     \end{tabular}
%     \caption{Training Time Comparison on LLaMA3.1-8B with the Stanford Alpaca Dataset.}
%     \label{additional_time}
% \end{table*}

\begin{table*}[ht]
    \centering
    \resizebox{0.95\textwidth}{!}{%
    \begin{tabular}{lcccccc}
        \toprule
        \multirow{2}{*}{Method} &
        \multicolumn{2}{c}{LoRA} &
        \multicolumn{2}{c}{DoRA} &
        \multicolumn{2}{c}{DyLoRA} \\
        \cmidrule(lr){2-3} \cmidrule(lr){4-5} \cmidrule(lr){6-7}
        & w/o Hyper & w/ Hyper & w/o Hyper & w/ Hyper & w/o Hyper & w/ Hyper \\
        \midrule
        Training Time (s) & 6893 & \textbf{5958} & 7525 & \textbf{6267} & 7129 & \textbf{6132} \\
        \bottomrule
    \end{tabular}}
    \caption{Training time comparison on LLaMA3.1-8B with the Stanford Alpaca dataset.}
    \label{additional_time}
\end{table*}

\begin{table*}[ht]
    \centering
    \resizebox{0.95\textwidth}{!}{%
    \begin{tabular}{@{}ccccccccc@{}}
        \toprule
        \multirow{2}{*}{Dataset} &
        \multicolumn{2}{c}{LoRA} &
        \multicolumn{2}{c}{AdaLoRA} &
        \multicolumn{2}{c}{DoRA} &
        \multicolumn{2}{c}{DyLoRA} \\
        \cmidrule(lr){2-3} \cmidrule(lr){4-5} \cmidrule(lr){6-7} \cmidrule(lr){8-9}
        & w/o Hyper & w/ Hyper & w/o Hyper & w/ Hyper & w/o Hyper & w/ Hyper & w/o Hyper & w/ Hyper \\
        \midrule
        GSM8K & 71.19 & 71.21 & 72.38 & 72.54 & 72.45 & 72.46 & 72.52 & 72.53 \\
        HumanEval & 42.26 & 42.27 & 43.54 & 43.56 & 43.51 & 43.53 & 43.43 & 43.45 \\
        \bottomrule
    \end{tabular}}
    \caption{Performance comparison of different methods on LLaMA3.1-8B (GSM8K and HumanEval).}
    \label{additional_performance}
\end{table*}

\subsubsection{Performance on Reasoning Benchmarks} We conduct a comprehensive evaluation of various  methods on the LLaMA3.1-8B model using two additional benchmark datasets, GSM8K and HumanEval. As demonstrated in the Table \ref{additional_performance}, our approach does not result in any performance degradation on these complex tasks.

\subsubsection{Convergence Time on Larger Models} We further evaluate our method on the larger Qwen2.5-14B model with training load benchmarks on the Stanford Alpaca and OpenPlatypus datasets. As shown in the Table \ref{14B}, our method achieves significantly faster convergence compared to the baselines.
\begin{table}[ht]
    \centering
    \begin{tabular}{ccc}
        \toprule
        Method & Stanford Alpaca & OpenPlatypus \\
        \midrule
        AdaLoRA & 15102 & 24857 \\
        \textbf{Ours} & \textbf{11553} & \textbf{18553} \\
        \bottomrule
    \end{tabular}
    \caption{Comparison of Training Time (s) on Qwen2.5-14B.}
    \label{14B}
\end{table}

\subsection{Hyperparameter Impact Analysis}
We investigate the sensitivity of HyperAdaLoRA's NLG performance to the orthogonality regularization coefficient $\gamma$. We finetune LLaMA3.1-8B with $\gamma$ values set to $\{0.1, 0.15, 0.2\}$. As shown in Table \ref{tab:gamma_sensitivity}, performance remains relatively stable across the tested $\gamma$ values. Although $\gamma = 0.2$ yields slightly better results in this specific setup, the differences are minimal. This indicates that HyperAdaLoRA is robust to variations in this hyperparameter, thereby simplifying its practical application.
\begin{table}[ht!] % Use [ht!] for better placement control
\small
\begin{tabular}{ccccc}
\toprule
\multirow{2}{*}{\textbf{\(\boldsymbol{\gamma}\)}} & \multicolumn{2}{c}{\textbf{Stanford Alpaca}} & \multicolumn{2}{c}{\textbf{Magpie}} \\
\cmidrule(lr){2-3} \cmidrule(lr){4-5}
 & BLEU-4 & ROUGE-1 & BLEU-4 & ROUGE-1 \\
\midrule
0.10 & 55.10 & 58.58 & 70.73 & 56.78 \\ % Updated based on image
0.15 & 55.06 & 58.42 & 70.70 & 56.68 \\ % Updated based on image
0.20 & 55.12 & 58.30 & 70.72 & 56.78 \\ % Updated based on image
\bottomrule
\end{tabular}

\centering
\caption{Performance of HyperAdaLoRA with different values of \(\gamma\). We conduct experiments using the LLaMA3.1-8B model on the Stanford Alpaca and Magpie-Pro-300K-Filtered datasets.}
\label{tab:gamma_sensitivity}
\end{table}

\begin{table*}[htbp]
    \centering
    \small
    \begin{tabular}{@{}ccccccc@{}}
        \toprule
        \textbf{Model} & \textbf{Method} & \textbf{Stanford Alpaca} & \textbf{Magpie-Pro-300K-Filtered} & \textbf{OpenPlatypus} \\
        \midrule
        \multirow{2}{*}{LLaMA3.1-8B} & AdaLoRA & 8125 & 19600 & 11900 \\
        &  HyperAdaLoRA (ours) &  \textbf{6650} &   \textbf{15720} &  \textbf{9750} \\
        \cmidrule{1-5}
        \multirow{2}{*}{Qwen2.5-7B} & AdaLoRA & 4240 & 15000 & 6750 \\
        & HyperAdaLoRA (ours) & \textbf{3500} &   \textbf{11000} &  \textbf{5500} \\
        \bottomrule
    \end{tabular}%
    \caption{Comparison of total training time (in seconds) for AdaLoRA and HyperAdaLoRA across natural language generation tasks. Experiments are conducted using the LLaMA3.1-8B and Qwen2.5-7B models on the Stanford Alpaca, Magpie-Pro-300K-Filtered, and OpenPlatypus datasets. HyperAdaLoRA consistently demonstrates lower training times across these settings.}
    \label{tab:comparison}
\end{table*}

\begin{figure}[ht]
\centering
% \begin{subfigure}[t]{0.32\textwidth}
%     \centering
%     \includegraphics[width=\textwidth]{latex/figures/3_MNLIDeBERTa-v3-base.png}
%     \caption{MNLI}
%     \label{fig:MNLI_MLP}
% \end{subfigure}
% \hfill
\begin{subfigure}[t]{0.23\textwidth}
    \centering
    \includegraphics[width=\textwidth]{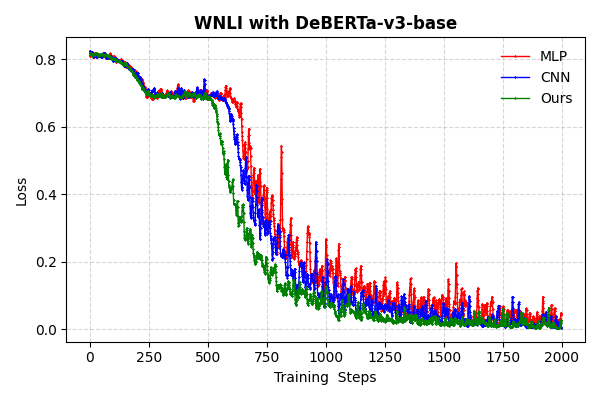}
    \caption{WNLI}
    \label{fig:WNLI_MLP}
\end{subfigure}
\hfill
\begin{subfigure}[t]{0.23\textwidth}
    \centering
    \includegraphics[width=\textwidth]{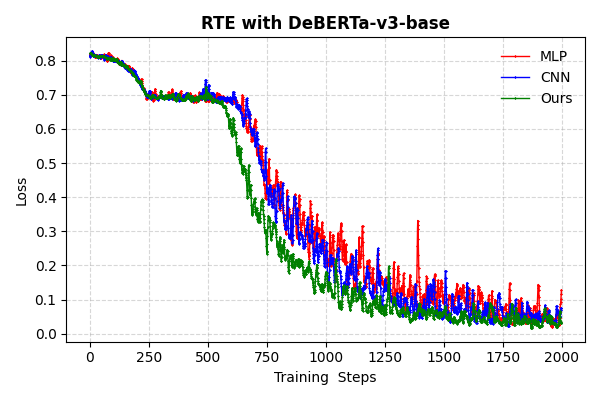}
    \caption{RTE}
    \label{fig:RTE_MLP}
\end{subfigure}
\caption{Comparison of training loss convergence for different HyperAdaLoRA hypernetwork architectures (MLP, CNN, BERT layer) on the DeBERTa-v3-base model for NLU tasks. The analysis is conducted across three natural language understanding tasks.}
\label{fig:hypernetwork_ablation}
\end{figure}

\begin{table*}[h]
    \centering
    \small
    \begin{tabular}{ccccc}
        \toprule
        \multirow{2}{*}{\textbf{Batch Size}} & \multicolumn{2}{c}{\textbf{AdaLoRA}} & \multicolumn{2}{c}{\textbf{HyperAdaLoRA (ours)}}  \\
        \cmidrule(lr){2-3} \cmidrule(lr){4-5}
        & Memory (MB) & Latency (ms / step) & Memory (MB) & Latency (ms / step) \\
        \midrule
        1 & 2758 & 123.16 & 2722 & 118.10 \\
        2 & 2798 & 132.60 & 2764 & 118.82  \\
        4 & 3232 & 149.38 & 3196 & 134.54  \\
        8 & 4115 & 189.39 & 4078 & 178.51  \\
        16 & 5906 & 284.33 & 5870 & 272.72  \\
        32 & 9600 & 486.16 & 9564 & 465.77  \\
        64 & 16566 & 882.55 & 16530 & 872.80  \\
        \bottomrule
    \end{tabular}
    \caption{Comparison of memory usage and training latency per step between AdaLoRA and HyperAdaLoRA across various batch sizes. The experimental settings are consistent with the implementation details described above.}
    \label{tab:efficiency}
\end{table*}

% \begin{table}[htbp]
%     \centering
%     \resizebox{\columnwidth}{!}{%
%     \begin{tabular}{@{}lccc@{}}
%         \toprule
%         & Stanford Alpaca & Magpie-Pro-300K-Filtered & OpenPlatypus \\
%         \midrule
%         \multirow{2}{*}{LLaMA3.1-8B} & 8125 & 19600 & 11900 \\
%         & 6650 & 15720 & 9750 \\
%         \cmidrule{1-4}
%         \multirow{2}{*}{Qwen2.5-7B} & 4240 & 15000 & 6750 \\
%         & 3500 & 11000 & 5500 \\
%         \bottomrule
%     \end{tabular}
%     }
%     \caption{Comparison of total training time (in seconds) between AdaLoRA and HyperAdaLoRA for natural language generation tasks. Each line is divided into two sublines. The upper one indicates the training time of AdaLoRA, while the lower one represents ours. Experiments are conducted using the LLaMA3.1-8B and Qwen2.5-7B models on the Stanford Alpaca, Magpie-Pro-300K-Filtered, and OpenPlatypus datasets. HyperAdaLoRA consistently demonstrates lower training times across these settings.}
%     \label{tab:comparison}
% \end{table}

\subsection{Ablation Study}

To demonstrate the contributions of our hypernetwork design, we compare the performance of HyperAdaLoRA with different hypernetwork architectures (MLP, CNN, and BERT layer) when finetuning DeBERTa-v3-base. Figure~\ref{fig:hypernetwork_ablation} shows the training loss curves of these three variants on the RTE and WNLI datasets. The results show that the choice of hypernetwork architecture affects the convergence speed. The BERT layer hypernetwork achieves the fastest convergence across all three datasets. This indicates that the attention mechanism is particularly effective at capturing the complex interdependencies among the elements of the $P$, $\Lambda$, and $Q$ matrices, thereby generating more efficient and targeted updates. In contrast, the MLP and CNN-based hypernetworks lag behind the BERT layer variant. The MLP, being the simplest architecture, shows the least acceleration, while the CNN provides intermediate results. This performance hierarchy is consistent with the representation capabilities of these architectures, further demonstrating the benefits of using complex mechanisms like attention to generate parameters in this context.

\subsection{Efficiency Analysis}
We analyze the computational load of our method compared to AdaLoRA. Table \ref{tab:efficiency} presents the GPU memory usage and per step training latency for both methods under different batch sizes. As shown in Table \ref{tab:efficiency}, HyperAdaLoRA exhibits a slight reduction in memory usage compared to AdaLoRA. Additionally, HyperAdaLoRA consistently demonstrates lower training latency per step. Although the per step reduction may appear modest, the faster convergence rate demonstrated earlier leads to a significantly shorter total training time to reach the target performance level. Therefore, HyperAdaLoRA achieves a notable improvement in training efficiency. This efficiency gain mainly stems from two factors: first, our hypernetwork is lightweight and adds minimal computational overhead; second, by generating task-specific parameters conditioned on context, it enables dynamic adaptation without iterative optimization, accelerating convergence. By modulating update direction and magnitude, it efficiently explores high-dimensional parameter spaces while preserving model expressiveness.  We provide an analysis of memory usage in Appendix~\ref{app:memory}.

% This improvement in efficiency can be primarily attributed to two key factors. First, the hypernetwork we employ is lightweight and does not introduce significant additional computational overhead. Second, the latency reduction benefits from the convergence acceleration enabled by the hypernetwork. By generating task-specific parameters conditioned on contextual inputs, hypernetworks enable dynamic adaptation without requiring explicit iterative optimization, effectively capturing complex parameter adaptation patterns in real time. Hypernetworks facilitate adaptive parameter updates by modulating both the update direction and magnitude according to the current model state. This dynamic modulation promotes efficient exploration of high-dimensional parameter spaces, accelerating convergence while preserving model expressiveness.

\section{Conclusion}
In this paper, we address the issue of slow convergence in AdaLoRA, an effective dynamic rank allocation method within the PEFT framework. To overcome this limitation, we propose HyperAdaLoRA, a novel and flexible framework that employs hypernetworks to dynamically generate the SVD-based low-rank parameters $(P, \Lambda, Q)$ central to AdaLoRA. By adopting an attention-enhanced hypernetwork architecture, HyperAdaLoRA captures fine-grained parameter dependencies and produces more targeted updates during training. This design enables the model to explore the optimization space more efficiently, leading to notably faster convergence compared to standard AdaLoRA training methods. Extensive experiments across multiple benchmarks and model scales show that HyperAdaLoRA achieves faster convergence while maintaining comparable or even better final performance. Furthermore, extension experiments on other LoRA-based methods clearly demonstrate the versatility and broad applicability of our approach.

\clearpage

\section*{Limitations}
In this work, we conduct extensive experiments to evaluate the effectiveness of our hypernetwork based training method in accelerating the training of various tasks. The results demonstrate that our approach can significantly speed up training without compromising performance. However, due to computational constraints, we have not yet been able to evaluate it on much larger models, such as those with 70 billion parameters. Exploring its scalability to models of this scale represents an important direction for future work. 
\bibliography{custom}

\appendix
\newpage

\section{Comprehensive Analysis of Hypernetworks for Accelerating Convergence}
\label{Comprehensive Analysis of Hypernetworks for Accelerating Convergence}
We present a more comprehensive theoretical analysis that elucidates how hypernetworks accelerate convergence.

\textbf{Context-Aware Task-Specific Parameter Generation for Accelerated Convergence:} Hypernetworks dynamically generate task-specific parameters based on contextual input information, enabling the direct output of optimal parameters or their low-rank representations for the current task. This approach circumvents the traditional, time-consuming explicit optimization process, thereby significantly reducing the number of training iterations. Such a mechanism is particularly well-suited for multi-task settings or scenarios with rapidly changing data distributions, allowing for rapid adaptation to new tasks without the need for parameter re-learning from scratch and effectively accelerating model convergence.

\textbf{Hypernetworks as Learnable Inner Optimizers for Dynamic Update Strategies:} Hypernetworks adjust the update direction and step size of parameters based on the current model state, effectively simulating a learnable inner optimizer. Under this mechanism, the model no longer relies on a fixed, uniform optimization strategy; instead, the hypernetwork determines how to update at each step, enabling a more flexible and efficient convergence trajectory. This approach inherits the principles of meta-learning while avoiding nested backpropagation, resulting in reduced computational cost and faster convergence.

\textbf{Structured Parameter Optimization and Stability Enhancement in High-Dimensional Spaces via Hypernetworks:} In high-dimensional parameter spaces, conventional optimization methods often suffer from getting trapped in local optima or oscillatory regions. Hypernetworks, by explicitly modeling the distribution of parameters, learn more structured update strategies that facilitate more efficient exploration of superior regions in the parameter space. In particular, by encoding contextual information as priors, hypernetworks provide more directed parameter generation schemes, thereby enhancing both the stability and efficiency of the training process.

\section{Details of Hypernetworks for LoRA, DoRA, and DyLoRA}
\label{appendix::detailsof}
In line with our formulation that extends AdaLoRA to HyperAdaLoRA in the paper, we now provide a parallel description of how hypernetworks are applied to LoRA, DoRA, and DyLoRA. For LoRA and DyLoRA, their weight update formula is the same as in standard LoRA: $\Delta W = BA$, where $B$ and $A$ are updated via backpropagation. We introduce two hypernetworks, $H_B$ and $H_A$, to generate the updated parameters: $B_{i+1} = H_B(B_i; \Phi_B)$, $A_{i+1} = H_A(A_i; \Phi_A)$, where $\Phi_B$ and $\Phi_A$ are the trainable parameters of the hypernetworks. $B_i$ and $A_i$ denote the parameters before the $i$-th iteration, and $B_{i+1}$ and $A_{i+1}$ denote the parameters after the update. For DoRA, the original method factorizes the LoRA low-rank update into a direction component and a scalar component. Let the directional part be $\Delta W_{\mathrm{dir}} = B_{\mathrm{dir}} A_{\mathrm{dir}}$, and let the scalar be $s$. The weight update can then be written as $\Delta W = s \cdot \Delta W_{\mathrm{dir}} = s \cdot B_{\mathrm{dir}} A_{\mathrm{dir}}$. In standard DoRA, $B_{\mathrm{dir}}$ and $A_{\mathrm{dir}}$ are directly updated via backpropagation. Analogously, we introduce hypernetworks for the directional component in DoRA. In particular, we use two hypernetworks, $H_{B_{\mathrm{dir}}}$ and $H_{A_{\mathrm{dir}}}$, to generate the updated parameters: $B_{\mathrm{dir}, i+1} = H_{B_{\mathrm{dir}}}(B_{\mathrm{dir}, i}; \Phi_{B_{\mathrm{dir}}})$, $A_{\mathrm{dir}, i+1} = H_{A_{\mathrm{dir}}}(A_{\mathrm{dir}, i}; \Phi_{A_{\mathrm{dir}}})$, where $\Phi_{B_{\mathrm{dir}}}$ and $\Phi_{A_{\mathrm{dir}}}$ are the trainable parameters of the hypernetworks.

\section{Processing Details of Parameter Matrices (e.g., $P$)}
\label{Processing Details of Parameter Matrices}
We describe the processing pipeline of the parameter matrix \(P\). Initially, \(P\) is sampled from a Gaussian distribution and treated as a temporary variable. At each training step, the hypernetwork \(H_P\) takes \(P\) as input and produces an updated version of \(P\), which is then used in the forward pass. During backpropagation, gradients are propagated through \(P\) to update the parameters of \(H_P\). This process is iteratively repeated throughout the training procedure.

\section{Implementation Details of the Pruning Strategy}
\label{Implementation Details of the Pruning Strategy}
During the fine-tuning process, every \(\Delta T\) steps, we perform a pruning operation over the singular value sets \(\Lambda\) of the LoRA weight matrices. For each LoRA parameter matrix \(W_i\), we compute its current gradient \(\nabla W_i\) and obtain its singular values \(\{\sigma_{ij}\}_{j=1}^r\), where \(\sigma_{ij}\) denotes the \(j\)-th singular value of the \(i\)-th matrix. To quantify the relative importance of each singular direction, we define an importance score as
\begin{equation}
    s_{ij} = |\sigma_{ij} \cdot \nabla \sigma_{ij}|
\end{equation}
which reflects both the magnitude of the singular value and its sensitivity to the gradient, indicating its contribution to parameter adaptation. After computing all scores, we identify the \(k\) smallest \(s_{ij}\) across all matrices and prune their associated singular values. This targeted pruning reduces the rank allocation for less significant directions and enables dynamic redistribution of the global rank budget, prioritizing more impactful components and improving overall parameter efficiency during training. Our parameter \(k\) is fixed and predefined. It is worth noting that the pruning strategy is not the primary focus or core contribution of our work. It is directly inherited from the original design of AdaLoRA and is used solely to maintain consistency and fairness with prior work, rather than to introduce a new pruning technique.

\section{Other Training Details}
\label{Training Details}
During training, we employ a linear warm-up strategy to smoothly ramp up the learning rate, thereby enhancing training stability and convergence. Specifically, the learning rate is linearly increased from zero to the preset maximum value over the first 500 steps, after which the main learning rate scheduler (cosine annealing) takes over. Model parameters are initialized with a normal distribution. We set the average budget per weight to three, giving a total budget equal to three times the number of weights, and use a pruning interval of 100 steps. To ensure a fair comparison, all these training configurations are kept consistent.

\section{$P\Lambda Q$ Parameter Similarity Analysis}
We conduct an average similarity analysis between the $(P \Lambda Q)$ parameters generated by our method and those produced by AdaLoRA. As shown in Table~\ref{tab:parameter-similarity}, the results indicate that the similarity between corresponding parameters consistently exceeds 0.98, demonstrating that the two methods are highly consistent at the parameter level and essentially equivalent in terms of performance. This finding further supports that our method can achieve comparable parameter representations while maintaining strong performance.
\begin{table}[h]
\centering
\begin{tabular}{c|ccc}
\toprule
Parameter & $P$ & $\Lambda$ & $Q$ \\
\midrule
Cosine Similarity & 0.9861 & 0.9878 & 0.9811 \\
\bottomrule
\end{tabular}
\caption{Cosine similarity between the $(P \Lambda Q)$ parameters of our method and AdaLoRA.}
\label{tab:parameter-similarity}
\end{table}

\section{Additional Results on Training Time and Performance}
\label{app:more_results}
We further present a comparative performance evaluation on the OpenPlatypus dataset, as well as a {training time comparison of LLaMA3.1-8B} on GSM8K and HumanEval. The corresponding results are summarized in Table~\ref{tab:openplatypus-results} and~\ref{tab:training-time}. These results demonstrate that our method improves training efficiency without compromising performance.

\begin{table}[htbp]
\centering
\resizebox{0.48\textwidth}{!}{
\begin{tabular}{c| c c c}
\toprule
\textbf{Model} & \textbf{Method} & \textbf{BLEU-4} & \textbf{ROUGE-1} \\
\midrule
\multirow{2}{*}{Qwen2.5-7B} 
  & AdaLoRA & 19.87 & 44.24 \\
  & Ours    & 19.92 & 44.65 \\
\midrule
\multirow{2}{*}{LLaMA3.1-8B}
  & AdaLoRA & 34.86 & 52.90 \\
  & Ours    & 35.95 & 53.00 \\
\bottomrule
\end{tabular}
}
\caption{Performance comparison of HyperAdaLoRA and AdaLoRA on OpenPlatypus.}
\label{tab:openplatypus-results}
\end{table}

\begin{table*}[t]
\centering
\resizebox{\textwidth}{!}{
\begin{tabular}{c|cccccccc}
\toprule
\textbf{Dataset} & 
\textbf{LoRA w/o} & \textbf{LoRA w/} & 
\textbf{AdaLoRA w/o} & \textbf{AdaLoRA w/} &
\textbf{DoRA w/o} & \textbf{DoRA w/} &
\textbf{DyLoRA w/o} & \textbf{DyLoRA w/} \\
\midrule
GSM8K     & 9863 & 8110 & 10542 & 9011 & 10892 & 9054 & 12305 & 10527 \\
HumanEval & 8625 & 7210 & 10281 & 8620 & 9532  & 8029 & 10085 & 8547  \\
\bottomrule
\end{tabular}
}
\caption{Comparison of training time (s) across methods with and without hypernetworks (w/o = without hypernetworks, w/ = with hypernetworks).}
\label{tab:training-time}
\end{table*}

\section{Comparison with the $\gamma$ = 0 Setting (without regularization)}
We evaluate our method under the $\gamma = 0$ setting on the LLaMA3.1-8B model using the Stanford Alpaca and Magpie datasets. As shown in Table~\ref{tab:alpaca-magpie}, setting $\gamma = 0$ leads to a slight, yet acceptable, drop in accuracy, while the training time remains largely unchanged. This indicates that our method can still achieve strong performance even without applying regularization.

\begin{table*}[htbp]
\centering
\begin{tabular}{c|cccc}
\toprule
\textbf{Method} & \textbf{Stanford Alpaca} & \textbf{Magpie} & \textbf{Stanford Alpaca (Time)} & \textbf{Magpie (Time)} \\
\midrule
Ours ($\gamma = 0$) & 54.88 & 70.36 & 6970 & 14920 \\
Ours             & 55.10 & 70.73 & 6650 & 15720 \\
\bottomrule
\end{tabular}
\caption{Evaluation of our method under $\gamma = 0$ on the Stanford Alpaca and Magpie datasets.}
\label{tab:alpaca-magpie}
\end{table*}

\section{Memory Analysis}
\label{app:memory}
We provide here an analysis of how our method reduces memory consumption. The number of parameters introduced by our hypernetwork is much smaller than the total number of fine-tuning parameters introduced in the baselines (e.g., $(P,\Lambda,Q)$ in AdaLoRA, or $(A,B)$ in LoRA). In these baseline methods, the optimizer needs to maintain separate first- and second-moment states for all these fine-tuning parameters, which consumes a considerable amount of memory. In our method, $(P,\Lambda,Q)$ / $(A,B)$ are generated on demand by a shared lightweight hypernetwork and are no longer treated as independent trainable parameters, so the optimizer only needs to maintain states for the hypernetwork parameters. This substantially reduces the amount of optimizer states, leading to a slightly lower peak memory footprint.

\section{Training Stability of Our Method}
Our proposed architecture is inherently stable to train for several reasons: (1) Shared optimization objective: the hypernetwork does not alter the main model’s optimization objective but merely imposes a structured parametrization on the LoRA parameters, so the system still minimizes the same task loss instead of engaging in an adversarial game; (2) Low-dimensional update space: the hypernetwork only operates on low-rank LoRA factors, which imposes strong constraints on the parameter space and restricts updates to a structured low-dimensional subspace, leading to a smoother and more tractable optimization landscape; (3) Aligned gradient signals: the main model and the hypernetwork share the same task loss and gradient signal, so there are no competing objectives and their optimization directions are statistically aligned; (4) Lightweight hypernetwork design: the hypernetwork is relatively small and trained with conservative optimization hyperparameters, behaving more like a lightweight adaptation module rather than a second large backbone; (5) LoRA bounded update magnitude: the low-rank structure of LoRA naturally bounds the magnitude of updates and gradient norms, so even when local gradients are large, the effective update space is constrained by the rank and bottleneck dimensions, which in practice helps prevent gradient explosion and numerical divergence.

\section{Case Study}
\label{Case Study}
\begin{figure}[ht]
    \centering
    \includegraphics[width=0.85\columnwidth]{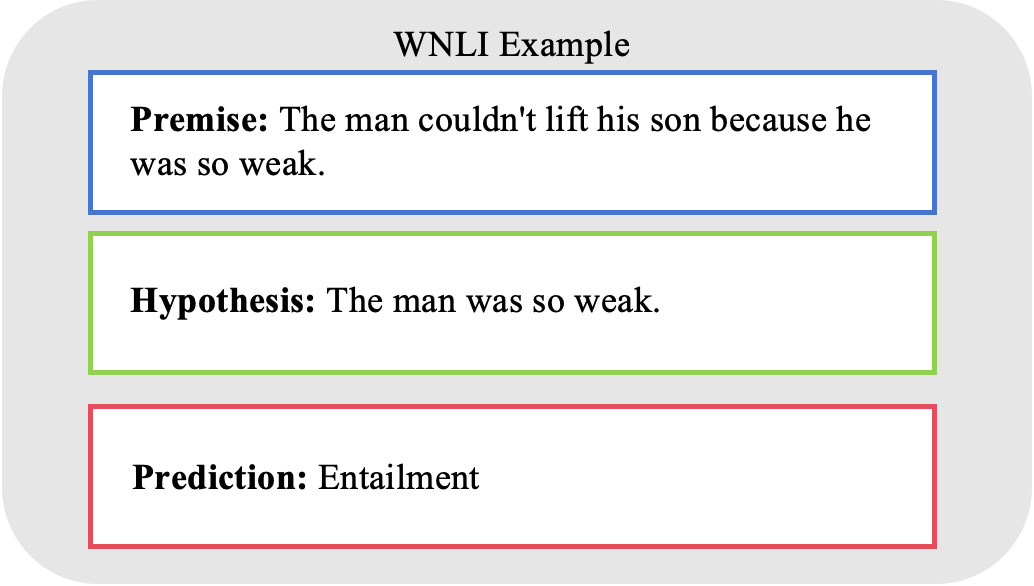}
    \caption{Examples generated by RoBERTa-base and DeBERTa-v3-base for the WNLI dataset.}
    \label{fig:WNLI}
\end{figure}

\begin{figure}[ht]
    \centering
    \includegraphics[width=0.85\columnwidth]{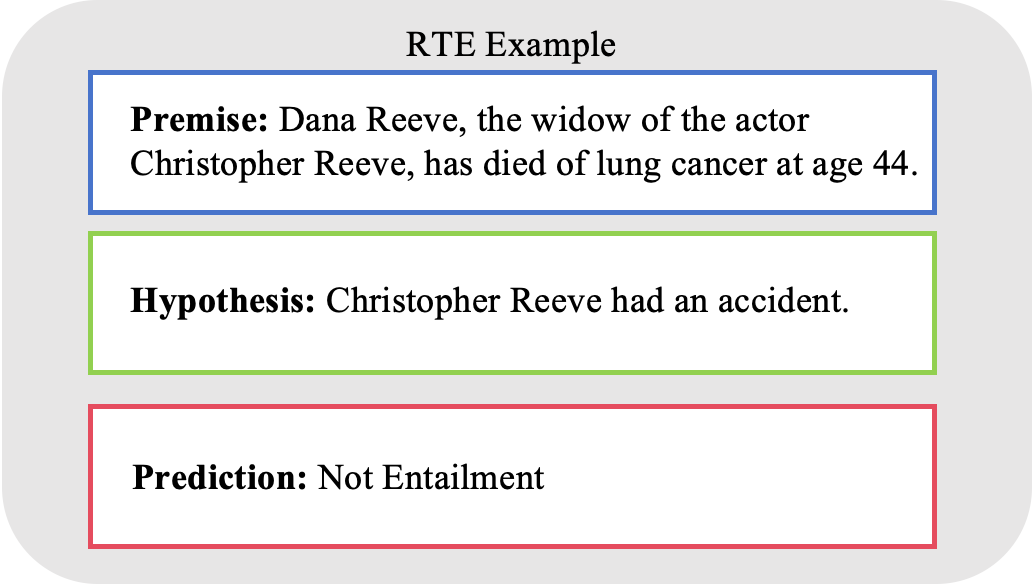}
    \caption{Examples generated by RoBERTa-base and DeBERTa-v3-base for the RTE dataset.}
    \label{fig:RTE}
\end{figure}

In Figure \ref{fig:WNLI} and \ref{fig:RTE}, we can see examples of two different natural language understanding tasks: WNLI and RTE. These examples demonstrate the models' capabilities in understanding and reasoning with textual information. These examples illustrate how models perform on different types of textual reasoning tasks when finetuned with our method. By fine-tuning RoBERTa-base and DeBERTa-v3-base models, we can enhance their performance on these tasks, thereby improving their ability to understand and reason with textual information.


\end{document}